\documentclass{lni}

\usepackage[bblfile=\jobname]{biblatex-readbbl}
\usepackage{booktabs}

\usepackage[]{blindtext}

\usepackage{subcaption}
\usepackage{graphicx}

\usepackage{tabularx}
\newcolumntype{K}[1]{>{\raggedright\arraybackslash}m{#1}} % linksbündig mit Breitenangabe
\newcolumntype{X}[1]{>{\centering\arraybackslash}m{#1}} % zentriert mit Breitenangabe
\newcolumntype{E}[1]{>{\raggedleft\arraybackslash}m{#1}} 

\begin{document}
%%% Mehrere Autoren werden durch \and voneinander getrennt.
%%% Die Fußnote enthält die Adresse sowie eine E-Mail-Adresse.
%%% Das optionale Argument (sofern angegeben) wird für die Kopfzeile verwendet.
\title[AutoMeet]{AutoMeet: a proof-of-concept study of genAI to automate meetings in automotive engineering}
%%%\subtitle{Untertitel / Subtitle} % falls benötigt
\author[1,2]{Simon Baeuerle}{}{}%
\author[2,3]{Max Radyschevski}{}{}%
\author[3]{Ulrike Pado}%

% Only display email & orcid once
\makeatletter
  \csgdef{@emailsandorcids1}{%  
 e-mail: \email{simon.baeuerle@kit.edu}%
%  \orcid{0000-0003-0296-7074}%
}
\makeatother
\makeatletter
  \csgdef{@emailsandorcids3}{%  
 e-mail: \email{ulrike.pado@hft-stuttgart.de}%
%  \orcid{0000-0003-0296-7074}%
}
\makeatother

\affil[1]{Institute of Automation and Applied Informatics (IAI), Karlsruhe Institute of Technology (KIT), 76344 Eggenstein-Leopoldshafen}%
\affil[2]{Mobility Electronics, Robert Bosch GmbH, 72770 Reutlingen}%
\affil[3]{Stuttgart University of Applied Sciences, 70174 Stuttgart}%
\maketitle

\begin{abstract}
%(70-150 Wörter)

In large organisations, knowledge is mainly shared in meetings, which takes up significant amounts of work time.
Additionally, frequent in-person meetings produce inconsistent documentation -- official minutes, personal notes, presentations may or may not exist. Shared information therefore becomes hard to retrieve outside of the meeting, necessitating lengthy updates and high-frequency meeting schedules.

Generative Artificial Intelligence (genAI) models like Large Language Models (LLMs) exhibit an impressive performance on spoken and written language processing. This motivates a practical usage of genAI for knowledge management in engineering departments: using genAI for transcribing meetings and integrating heterogeneous additional information sources into an easily usable format for ad-hoc searches.

We implement an end-to-end pipeline to automate the entire meeting documentation workflow in a proof-of-concept state: meetings are recorded and minutes are created by genAI. These are further made easily searchable through a chatbot interface. The core of our work is to test this genAI-based software tooling in a real-world engineering department and collect extensive survey data on both ethical and technical aspects. Direct feedback from this real-world setup points out both opportunities and risks:
a) users agree that the effort for meetings could be significantly reduced with the help of genAI models, 
b) technical aspects are largely solved already,
c) organizational aspects are crucial for a successful ethical usage of such a system.

\end{abstract}


%%%%%%%%%%%%%%%%%%%%%%%%%%%%%%%%%%%%%%%%%%%%%%%%%%%%%%%%%%%%%%%%%%%%%%%%%%%%%%%%%%%%%%%%%%%
%%%%%%%%%%%%%%%%%%%%%%%%%%%%%%%%%%%%%%%%%%%%%%%%%%%%%%%%%%%%%%%%%%%%%%%%%%%%%%%%%%%%%%%%%%%
\section{Introduction}
In many work environments, especially in large companies, meetings are a primary means for spreading and updating information. This is a high effort, low transparency strategy which may actually delay information flow. Especially when conducted online, they are tiring to the participants \cite{riedl_stress_2022}. 

Digitalization of the information discussed in meetings with the use of generative AI is an obvious solution given the power of Large Language Models~\cite{Bommasani2021FoundationModels}. However, \cite{Tkalac2024} in an overview study of digital internal communication in organizations point out that social aspects of digitalization and the effectiveness of tool use are generally not in focus. Instead, studies often center on technical aspects of digitalizing internal communication. However, during technical method development also, there is little engagement with potential users according to a meta-study on Textual Summarization research \cite{liu-etal-2023-responsible}. Therefore, we identify user feedback and user involvement as an important area of inquiry.

Specifically looking at the use of generative AI (genAI), an obvious area of (user) concern comes into focus. In a case study of strategies for offering generative AI-enhanced products to customers, \cite{HakiEtAl2025} recommend, among other things, a careful implementation of data privacy safeguards, regulating the access to genAI and the control over its input and output. \cite{SöllnerEtAl2025} also identify concerns for the accuracy of genAI output and the possible presence of bias. Taken together, these issues ultimately affect user acceptance. Indeed, in a 2024 survey study \cite{Lünendonk2024}, 47\% of responses state a lack of user acceptance as a significant challenge for the use of genAI in their company.

Therefore, to be able to harness the power of current genAI techniques for meeting summarization in a way that satisfies users, we set out to ask what users want and need from an effective digital tool that they feel comfortable to use, in order to be able to ultimately provide a matching implementation. 

We study the automation of the end-to-end meeting workflow as opposed to solely the summarization of individual meetings, and implement proof-of-concept tooling as a demo to potential future users. On the basis of this demo, we gather feedback via a survey which provides valuable insights with respect to the practical use of genAI models.

% methodisches Beispiel-Paper
% https://link.springer.com/article/10.1365/s40702-025-01179-3

%%%%%%%%%%%%%%%%%%%%%%%%%%%%%%%%%%%%%%%%%%%%%%%%%%%%%%%%%%%%%%%%%%%%%%%%%%%%%%%%%%%%%%%%%%%
%%%%%%%%%%%%%%%%%%%%%%%%%%%%%%%%%%%%%%%%%%%%%%%%%%%%%%%%%%%%%%%%%%%%%%%%%%%%%%%%%%%%%%%%%%%
\section{State of the art in Meeting Summarization}

Meeting summarization is no trivial task, given that several speakers are present and may talk over one another, that topics may drift and non-verbal cues can be present~\cite{ijcai2022}.

Technically, the task is often broken down into the distinct steps of transcription and summarization. Transcription transforms speech recordings into text transcripts (text-to-speech, TTS). It is one field that demonstrates the large advantages of neural network approaches on natural language processing, with performance gains of over 50\% relative to previously used techniques \citep{PrabhavalkarEtAl24}. At the same time, commodification is sufficiently advanced that sophisticated models like Whisper\footnote{\url{https://github.com/openai/whisper}} are released under open-source licenses and can be run on end-user hardware without access to external servers. 
Not surprisingly, meeting software such as Microsoft Teams\footnote{\url{https://support.microsoft.com/en-us/office/transcribe-your-recordings-7fc2efec-245e-45f0-b053-2a97531ecf57}} or Zoom\footnote{\url{https://support.zoom.com/hc/en/article?id=zm_kb&sysparm_article=KB0064927}} also offer a transcription feature for recorded meetings.

The summarization task starts with meeting transcripts and has the goal of "condens[ing] the original dialogue into a shorter version covering salient
information"~\citep{FengEtAl2022}. Specifically for meeting summaries, \citep{FengEtAl2022} survey summarization research and find approaches that address the treatment of non-literal speech and non-verbal cues, as well as strategies for handling very long summaries. \cite{rennard-etal-2023-abstractive} in their own survey add to this the challenges of redundancy and incompleteness found in spoken language and the heterogeneity of the meeting domain in general, where meetings may be partly or fully dominated by a presentation by one speaker, or by multi-way discussions. 

Summaries were usually created using fine-tuned deep-learning models, but a recent trend is to prompt Large Language Models (LLMs) and to exploit their ability to semantically represent their input and create high-quality text output \citep{rennard-etal-2023-abstractive}. In a comparative study by \cite{laskar-etal-2023-building} on summarization performance of prompted LLMs, the GPT models by OpenAI dominated the field; however, even smaller open-source models from the Llama family proved robust to use if cost or data privacy concerns do not allow the use of closed models.

Further afield, \cite{Schneider2025policies} present an approach to creating meeting summarization on-line, that is, during the meeting. This approach however makes it harder to integrate additional documents like presentation slides, which are often available only after the fact, and which are needed to create summaries that correctly take meeting context into account and are, ideally, personalized \citep{kirstein-etal-2024-tell}. On-line summarization also does not allow a manual minutes revision step by the meeting convener in order to verify accuracy and completeness and raise user trust in the AI-generated minutes. This step can only be completed after the meeting has finished. 

For real-world applications, commercial IT systems such as \textit{otter.ai} or \textit{fireflies.ai} are already available to summarize meetings; meeting providers such as \textit{Zoom} also offer automated meeting summarization\footnote{\url{https://support.zoom.com/hc/de/article?id=zm_kb&sysparm_article=KB0058013}} based on the spoken content of the meeting.

%
%%%%%%%%%%%%%%%%%%%%%%%%%%%%%%%%%%%%%%%%%%%%%%%%%%%%%%%%%%%%%%%%%%%%%%%%%%%%%%%%%%%%%%%%%%%
%%%%%%%%%%%%%%%%%%%%%%%%%%%%%%%%%%%%%%%%%%%%%%%%%%%%%%%%%%%%%%%%%%%%%%%%%%%%%%%%%%%%%%%%%%%
\section{Proposed end-to-end pipeline}

\begin{figure}[!ht]
  \centering
  \includegraphics[width=.95\textwidth]{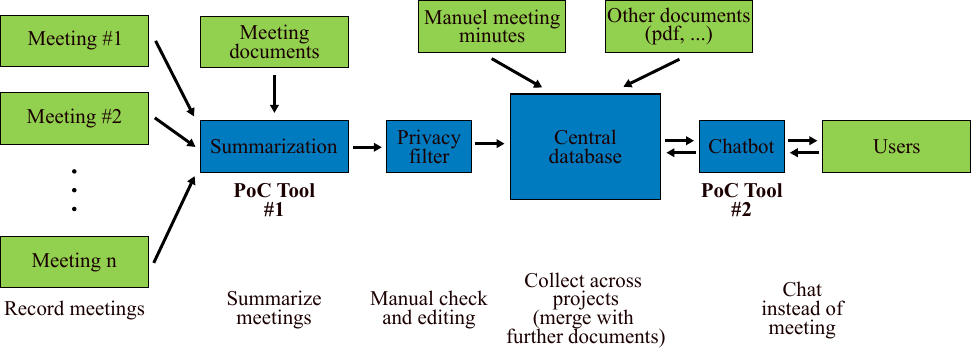}
  \caption{Envisioned end-to-end \textit{AutoMeet} pipeline for an automated meeting workflow supported by genAI. To cover the functionality in a proof-of-concept (PoC) state, we implement two standalone PoC tools: the summarization tool and the chatbot tool.}
  \label{fig:03_full_pipeline}
\end{figure}

Figure~\ref{fig:03_full_pipeline} shows a full information management system: meeting recordings are automatically transcribed and, together with additional documents like presentation slides, are processed into individual meeting summaries in the \textit{Summarization} module. A prompted LLM is used to summarize the technical content of both while automatically excluding personal data. This core module is implemented as one of the two proof-of-concept tools to demonstrate this functionality on actual meetings. 

The \textit{Privacy filter} module contains a manual check of the resulting summary. This includes both the removal of potentially remaining personal data and a correction of the technical content. The corrected meeting minutes are saved to a database along with relevant metadata such as the meeting participants, an identifier of the project and the persons or groups who may access the uploaded meeting minutes. Manual meeting minutes and other supplementary documents can be added to the database in the same way. 

The second core module, the \textit{chatbot}, provides easy text-based access to the compiled meeting information. Using the Retrieval-Augmented Generation strategy \cite{LewisEtAl2020}, documents relevant to the user query are retrieved and forwarded to the LLM. In this way, the system answer is constructed from reliable information sources rather than the model's implicit information. For example, the user could request information about the status of a project, and the LLM will generate an answer based on all relevant documents which the current user is allowed to access. The documents which are used by the LLM to generate the answer are displayed along with the respective metadata, to inform the user about the underlying sources of information and their time stamp. This way, the user can identify outdated information and see which statements were made by the most trustworthy contributors, e.g., domain experts. 

We see the implementation and real-life evaluation of the full pipeline as a potential mid- to long-term target. In the scope of this study, we implement only the two central parts of the pipeline in order to study the real-world effects in an early phase: the summarization step and the chatbot (with access to summarized meetings). Those two proof-of-concept tools are presented in sub-section~\ref{sec:32_poc_tools}. The tools are then introduced to actual development engineers and their feedback is collected via a survey. This survey aims to capture the organizational aspects of the end-to-end concept. It specifically includes questions on aspects, that would either prevent or enable developers to use the LLM tooling in daily work in a sustainable and acceptable way.

\subsection{Proof-of-concept tooling}\label{sec:32_poc_tools}

% Used models: Whisper, LLama, GPT4o..

\begin{figure}
  \centering
  \includegraphics[width=.95\textwidth]{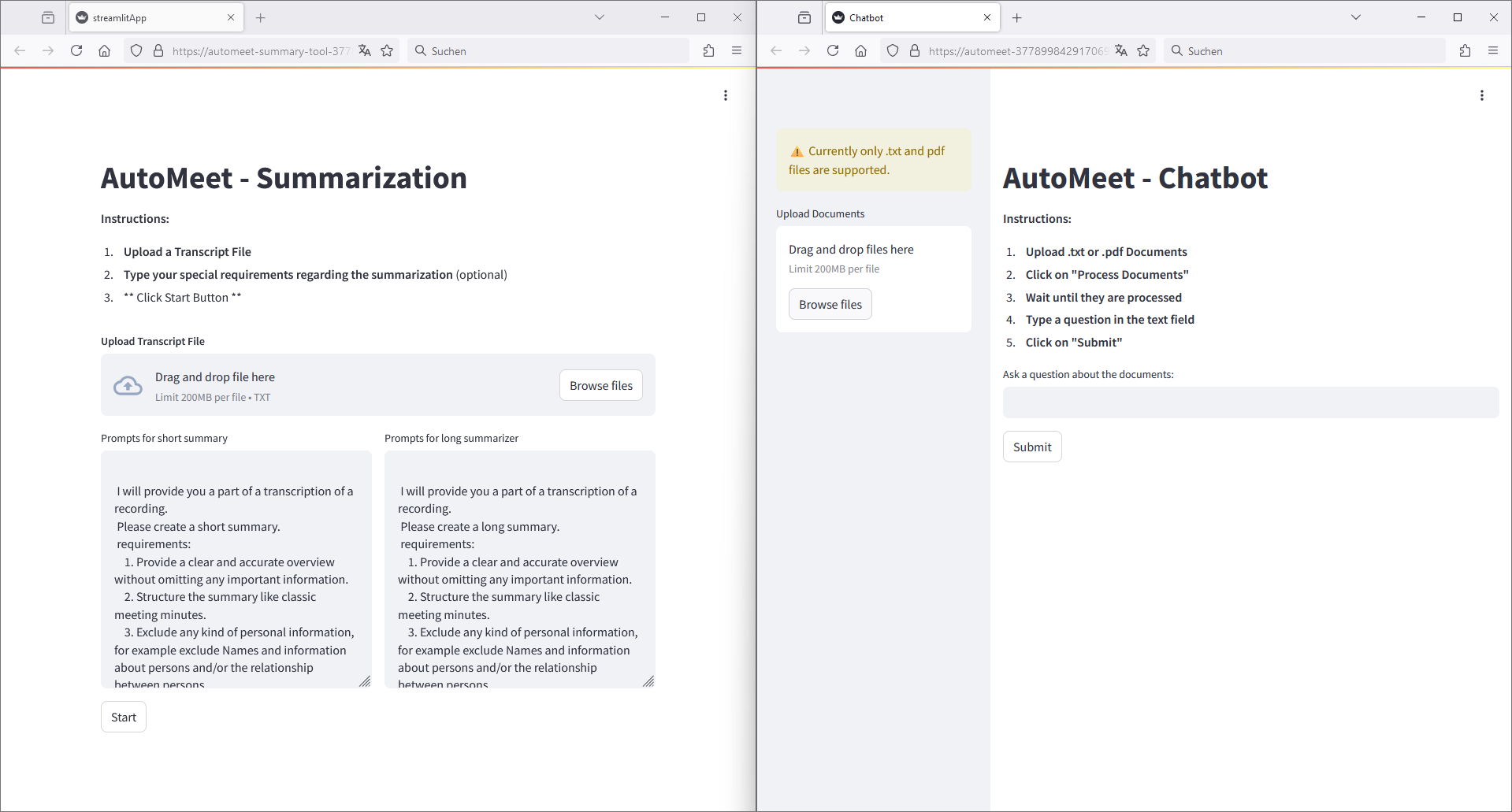}
  \caption{The User Interface of both proof-of-concept (PoC) tools that we used during our study. Left: summarization tool. Right: chatbot tool.}
  \label{fig:03_poc_tools}
\end{figure}

We first describe the first proof-of-concept tool with the minimum functionality needed to automatically summarize meeting contents.
We then go on to outline the second proof-of-concept tool which allows a direct interaction with those summarizations across multiple projects.
The documents are organized by the respective user: instead of using a full-scale database system during our proof-of-concept tests, each user manages the relevant documents on a local folder on his personal device.
Figure~\ref{fig:03_poc_tools} shows screenshots of the user interfaces of both tools. 
Both tools rely on the GPT-4o LLM model~\footnote{\url{www.openai.com}} and are implemented using the LangChain library~\footnote{\url{www.langchain.org}}.

The summarization tool takes a meeting recording created by the organisation's online meeting tool. It transcribes the recording's audio track outputs using the Whisper~\footnote{\url{https://github.com/openai/whisper}} open-source software package, deployed locally for data protection. The Word Error Rate (where 0 is best and 1 is 100\% error) for the transcriptions ranges between .23 for the best case of a single-speaker presentation recorded in good quality to 0.65 for a mix of a single-speaker presentation and multi-speaker discussion. We will discuss the quality of the transcriptions below as users comment on them.

From the transcript, the GPT-4o LLM creates summaries. The organisation's custom OpenAI API interface ensures data protection. The LLM was prompted to omit personal data like names from the transcript. The summarization tool produces both a short summary which is easier to edit and a long summary which contains more information. To achieve this, we use an iterative refinement technique~\footnote{As described by the LangChain development team at \url{https://python.langchain.com/docs/how_to/summarize_refine/}} to process long transcripts.

The next step in the processing pipeline (cf. Fig.~\ref{fig:03_full_pipeline}) is privacy filtering. The summaries were checked manually for private data, but none was found in our test cases involving more than 20 meeting summaries. The prompt for the summarizing LLM appears to have been sufficient on its own, so the explicit filtering step can be reduced to an editing of the technical contents.

The chatbot tool~\footnote{based on \{url{https://github.com/vikrambhat2/RAG-Implementation-with-ConversationUI.git}} uses a Retrieval Augmented Generation (RAG) pipeline \cite{LewisEtAl2020} to search for relevant meeting summaries appends them to the LLM prompt. The chatbot also uses previous conversations as relevant history. It cannot yet access a unified database as envisioned in Fig.~\ref{fig:03_full_pipeline}; instead relevant documents and collections of documents were made available manually for demo and tests.

\paragraph{Expert Ratings}
%Expertenbewertung auf Skala 1-6 \\
%Reference to related work with professional implementation, e.g. Schneider2025 --> PoC tools can be %exchanged for professional tools if needed (rouge metric)\\

%Hist_Q6.1
As genAI chatbots are used widely and survey participants have experience with genAI models like GPT4o in this context, we do not perform an in-depth validation of the chatbot tool.
However, summarization tools are not commonly used yet.
We validate the functionality of our summarization tool by asking trial users for a performance rating.
The rating for the quality of the short and long summaries from our PoC tooling is assessed on a scale from 1 (very good) to 6 (very bad), resulting to 2.58 on average (StdDev 1.32, n=12).
From full-text replies, survey participants liked the good structure and clear action points.
Reported drawbacks are a limited quality of the summaries created by the GPT4o LLM model, which is not comparable to the quality generated by human experts.
Furthermore, special vocabulary is not recognized by the current transcription model, e.g. \textquotedblleft centering\textquotedblright ~instead of \textquotedblleft sintering\textquotedblright.
This tool serves only to support our study.
For a stable deployment, we recommended to shift to a more stable tooling. %e.g. Schneider2025

%%%%%%%%%%%%%%%%%%%%%%%%%%%%%%%%%%%%%%%%%%%%%%%%%%%%%%%%%%%%%%%%%%%%%%%%%%%%%%%%%%%%%%%%%%%
%%%%%%%%%%%%%%%%%%%%%%%%%%%%%%%%%%%%%%%%%%%%%%%%%%%%%%%%%%%%%%%%%%%%%%%%%%%%%%%%%%%%%%%%%%%
\section{User Needs and Feedback}
Both the end-to-end \textit{AutoMeet} pipeline and the two PoC tools are presented to developers from different automotive R\&D departments. The concept and the functionality are explained. The developers can access the PoC tools and test them on actual meetings. Afterwards, the feedback from this real-world setting is collected via a survey. The survey contains a total 23 questions. We received a total of 42 survey replies. The survey feedback is included even if some questions were not filled out. The number of responses is indicated for each question individually. The results are presented in the following subsections. 

\subsection{Current status of information exchange}
\begin{figure}[htb]
  \centering
  \includegraphics[width=.95\textwidth]{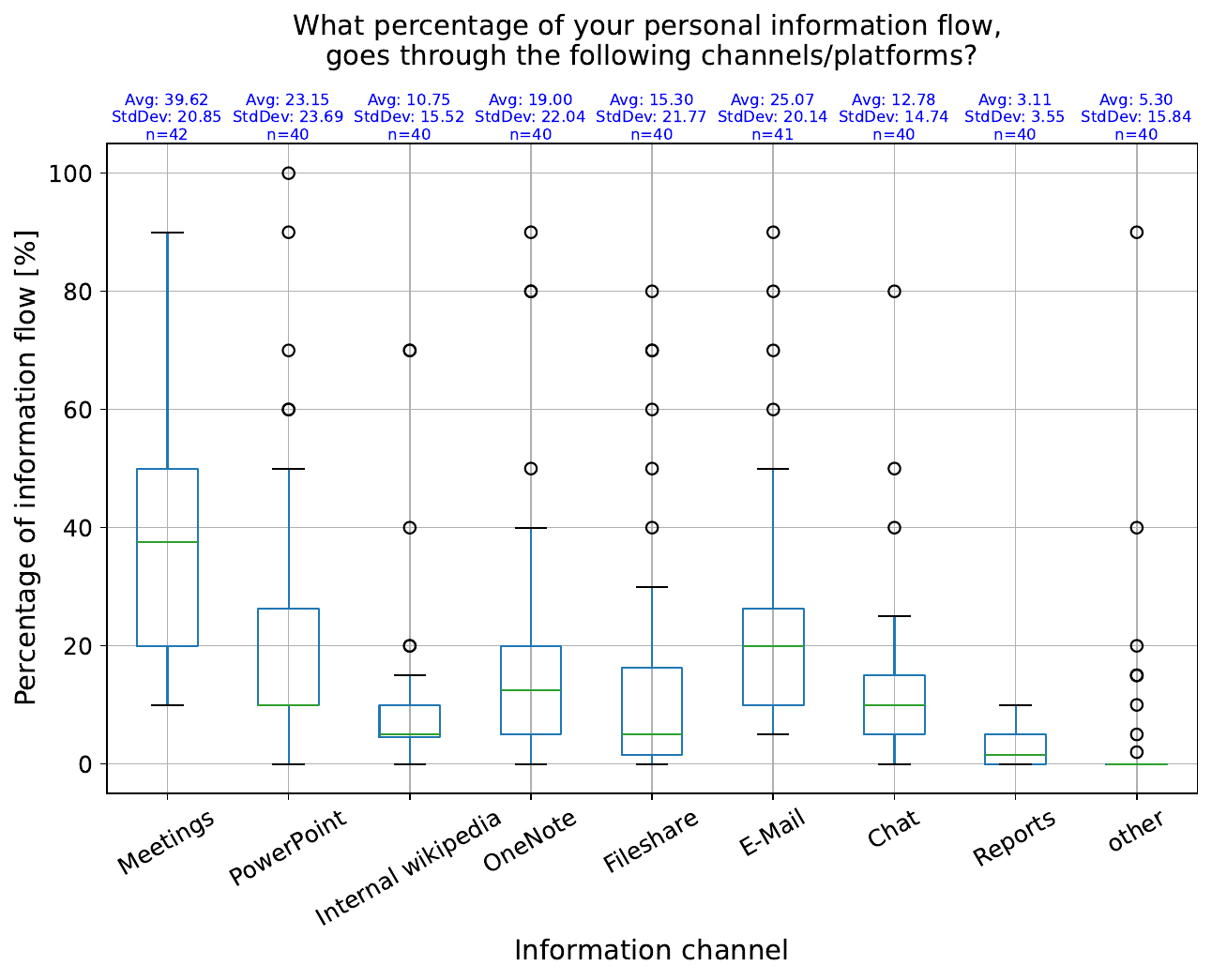}
  \caption{Reported proportion of successful information exchange for different information channels}\label{fig:04_information_channels}
\end{figure}

Figure~\ref{fig:04_information_channels} shows an overview of how much information is perceived to successfully flow through different channels. Please note that the channels overlap with each other and do not add up to 100\,\%: for example, power point slides may be shown during a meeting or shared via an email. This fraction of information flow will thus appear in both channels.

Effectively, the channels \textit{Meetings}, \textit{E-Mail} and \textit{MS PowerPoint} slides are mainly used. These main channels are entirely unstructured channels, with unclear guidelines regarding documentation and management. Meetings specifically are reported to make up for about 40\,\% of the information flow. More structured channels such as reports make up only about three percent of the information flow. The satisfaction with the current status of information exchange is reported to be 3.4 on average (StdDev 1.1, n=42) on a scale from 1 (very dissatisfied) to 6 (very satisfied).

\begin{figure}[htb]
  \centering
    \begin{subfigure}[b]{0.458\textwidth}
    \includegraphics[width=\textwidth]{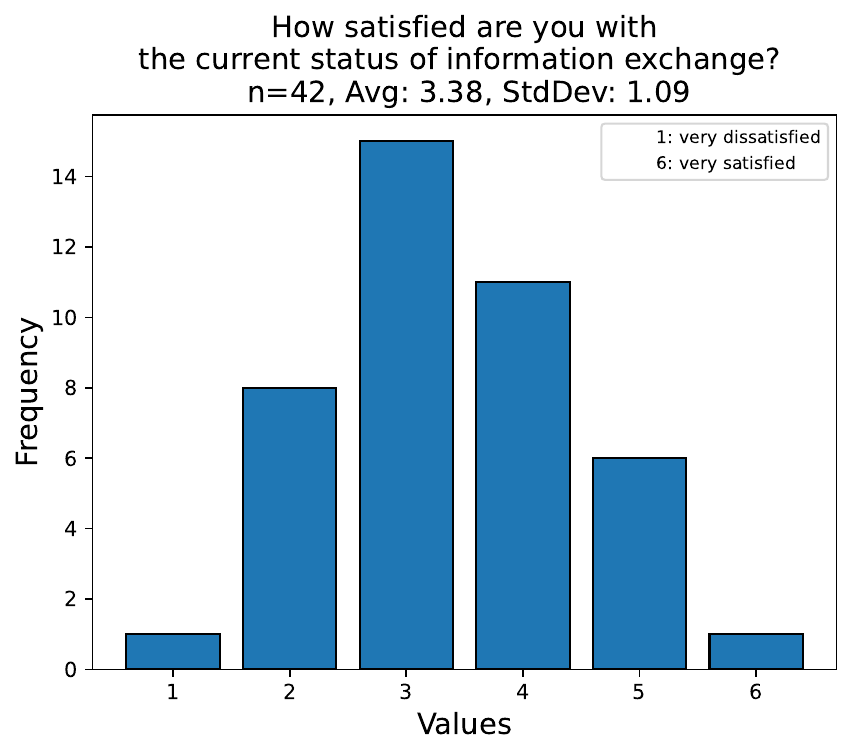}
    \end{subfigure}
    \begin{subfigure}[b]{0.458\textwidth}
    \includegraphics[width=\textwidth]{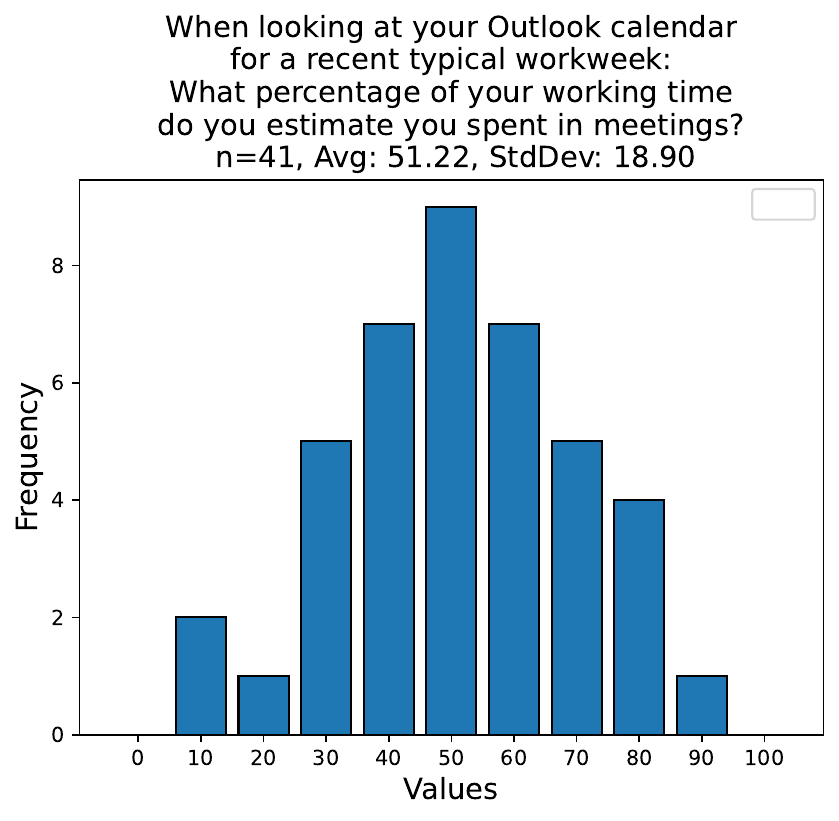}
    \end{subfigure}
  \caption{Status quo of information exchange: user satisfaction (left) and percentage of working time spent in meetings (right)}\label{fig:04_statusquo_exchange}
\end{figure}

\begin{figure}[htb]
  \centering
    \begin{subfigure}[b]{0.458\textwidth}
    \includegraphics[width=\textwidth]{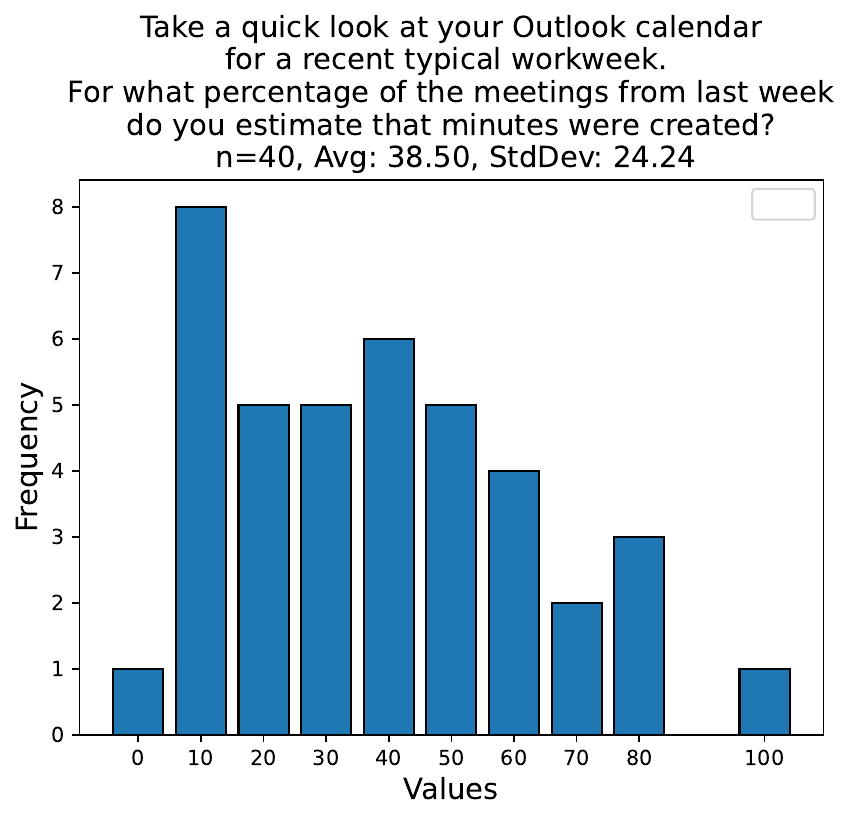}
    \end{subfigure}
    \begin{subfigure}[b]{0.458\textwidth}
    \includegraphics[width=\textwidth]{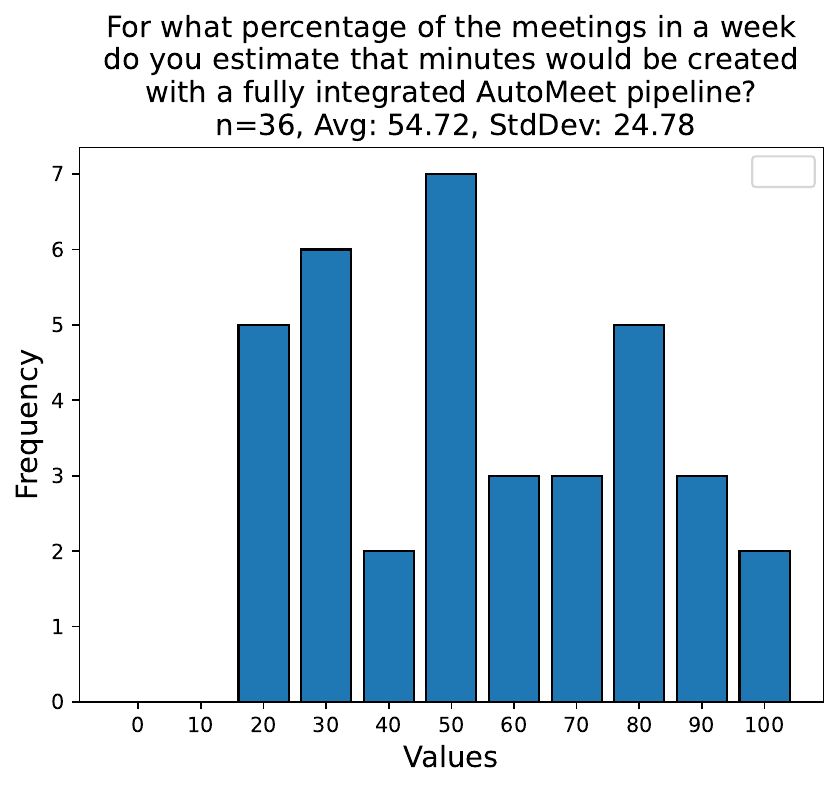}
    \end{subfigure}
  \caption{Proportion of meetings with created meeting minutes a) at the current state (left) and b) estimated with \textit{AutoMeet} (right)}\label{fig:04_minute_creation}
\end{figure}

\begin{figure}[htb]
  \centering
    \begin{subfigure}[b]{0.458\textwidth}
    \includegraphics[width=\textwidth]{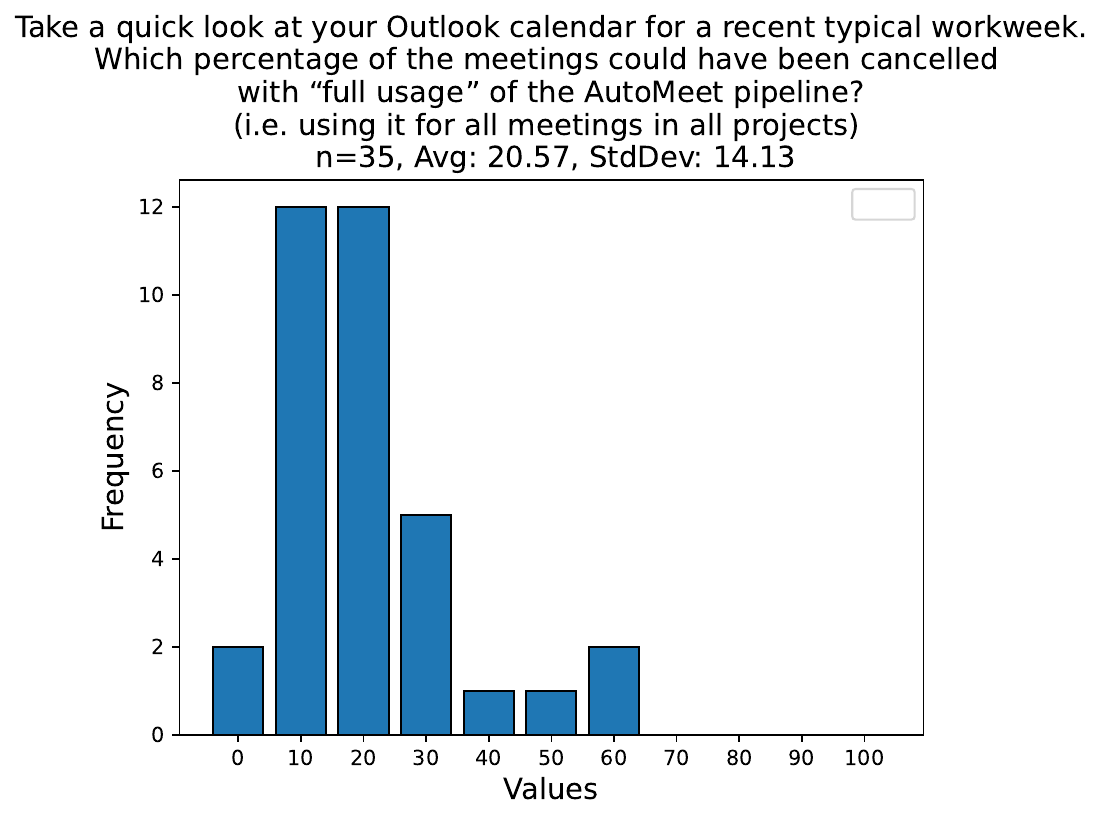}
    \end{subfigure}
    \begin{subfigure}[b]{0.35\textwidth}
    \includegraphics[width=\textwidth]{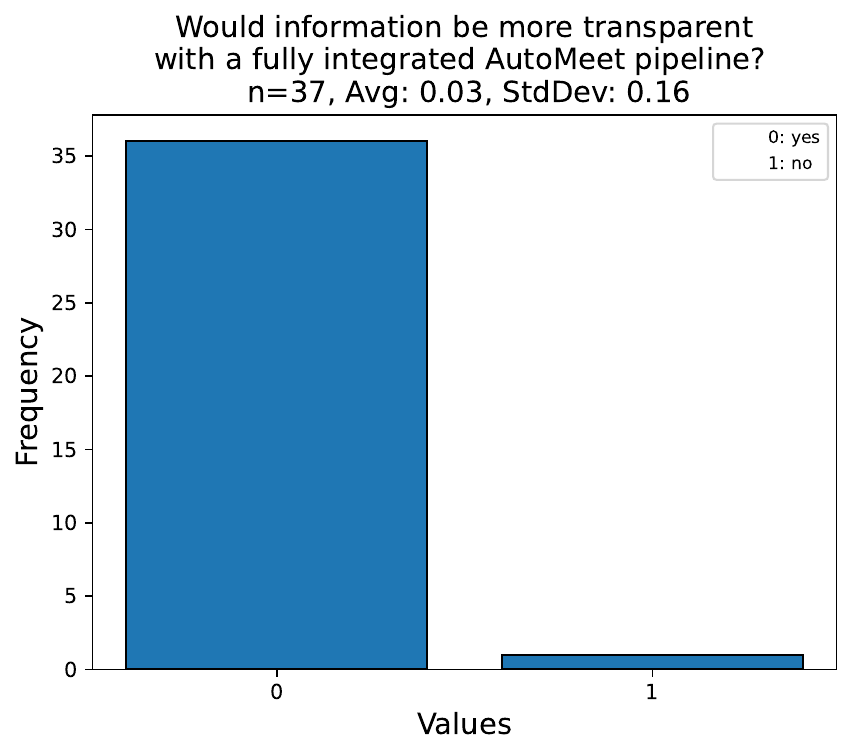}
    \end{subfigure}
  \caption{Benefits of an \textit{AutoMeet} pipeline with respect to meeting efforts (left) and information transparency (right) as estimated by survey participants}\label{fig:04_automeet_benefits}
\end{figure}

The percentage of meetings for which meeting minutes are created is currently estimated to be at 38.50\,\% (StdDev 24.24, n=40). 12 participants are satisfied with the current proportion of created minutes, whereas 28 are not satisfied.
With a fully integrated \textit{AutoMeet} pipeline, the proportion of created meeting minutes is estimated to raise from 38.50\,\% to 54.7\,\% on average (StdDev 24.78, n=36).
The distribution of created meeting minutes is visualized in Figure~\ref{fig:04_minute_creation} both for the current state and a potential future state with \textit{AutoMeet}.
Participants report to spend 51.2\,\% of their working time in meetings on average (StdDev 18.90, n=41).
The percentage of meetings that could have been cancelled with a full usage of the \textit{AutoMeet} pipeline (i.e. using it for all meetings in all projects) is estimated to be at 20.6\,\% on average (StdDev 14.13, n=35).
35 out of 36 participants responded that information would be more transparent with a fully integrated \textit{AutoMeet} pipeline.
The estimated benefits are visualized in Figure~\ref{fig:04_automeet_benefits}.

The feedback regarding the question \textit{In your opinion, what are the reasons why meeting minutes are not created and/or not used?} can be summarized as follows.
The main reason seems to be a lack of time (n=14), which coincides with a high effort to create (n=8).
Many state that it is difficult to simultaneously discuss and create meeting minutes (n=8) and that the responsibilities of who should create the minutes are not clearly defined (n=6).
Some assume that they will not be used anyway (n=4), doubt that the minutes can capture all relevant information (n=3). This may be due to a perceived poor quality of minutes (n=3). Few rely on the assumption, that actually important information will be repeated anyway during a follow-up meeting (n=2).

The feedback to this question overlaps with the question \textit{What would you change about the current status of information exchange?}. Reported numbers are aggregated for both questions in the current paragraph.
The following challenges regarding current documentation platforms are identified:
too many different information channels or platforms are used in parallel (n=13), with an unclear standardization of which information should be stored where (n=6).
The documentation is perceived to have poor quality (n=8).
Especially with regard to the documentation of meetings, meeting minutes seem to be sometimes missing entirely (n=4).
The following challenges regarding meetings themselves are reported as follows:
poor structure and preparation (n=5), parallel work of participants (n=2) and too many participants (n=2).

Based on the survey feedback for the average time spent in meetings and the proportion of meetings that could be cancelled with an \textit{AutoMeet} pipeline, savings can be estimated to be around 10.5\,\% of saved working time per employee. This saving of time efforts can be combined with hourly rates and the number of employees for individual departments or companies.

\subsection{Feedback to PoC tooling}
A core component of the \textit{AutoMeet} pipeline is the recording functionality.
Without any kind of recording, automated processing of the communicated information is not possible.
The willingness to record meetings is assessed with the survey question \textit{To automatically create meeting minutes, recording the meeting is necessary. The recording will be deleted after processing and will not be used for any other purposes. Would you agree to have your meetings recorded in this way/for documentation purposes?}. The replies are on average at 1.59 (StdDev 0.85, n=41) on a scale from 1 (yes) to 4 (no). The median reply is 1 (yes), which means a majority of users would agree to be recorded under those boundary conditions.

We summarize the replies to the question \textit{What reasons, in your opinion, personally speak against recording meetings?}.
The most prominent reason is the protection of data that is personal, sensitive or otherwise confidential (n=12).
Some replies express a more vague feeling of not being comfortable with being recorded (n=4).
Others clearly address concerns regarding premeditated misuse of recordings beyond the intended context (n=4).
Recordings are perceived to inhibit open discussion and block creative ideas (n=4).
Recordings also save statements in a fixed way, even though respective statements are not intended to be fixed (n=4).

\begin{figure}[htb]
  \centering
    \begin{subfigure}[b]{0.458\textwidth}
    \includegraphics[width=\textwidth]{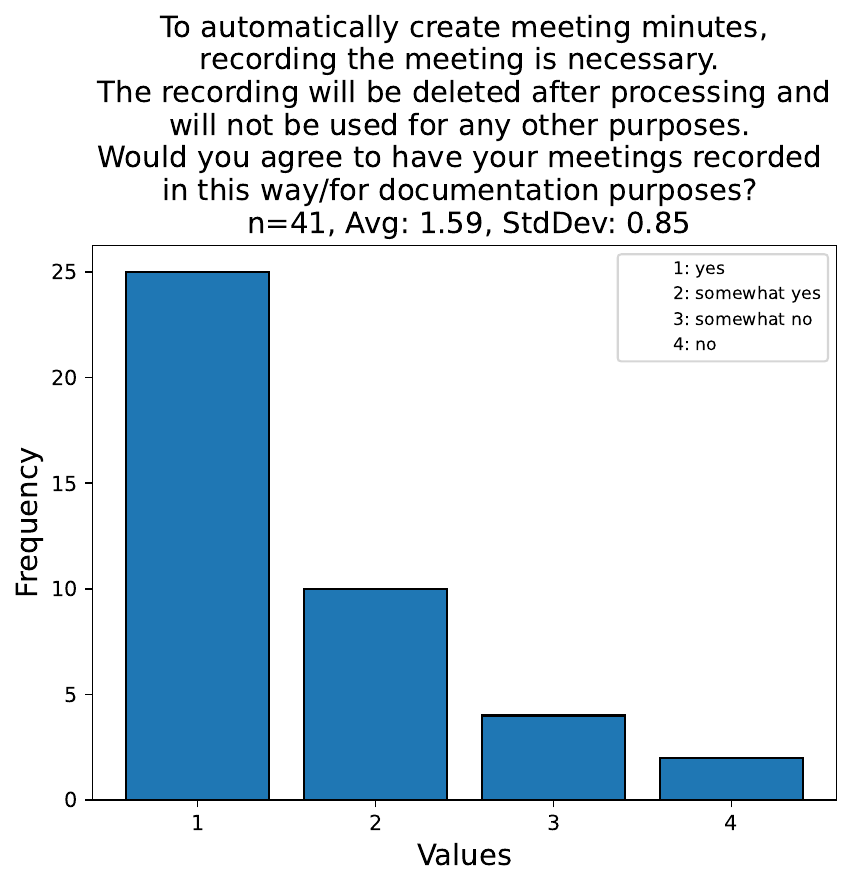}
    \end{subfigure}
    \begin{subfigure}[b]{0.458\textwidth}
    \includegraphics[width=\textwidth]{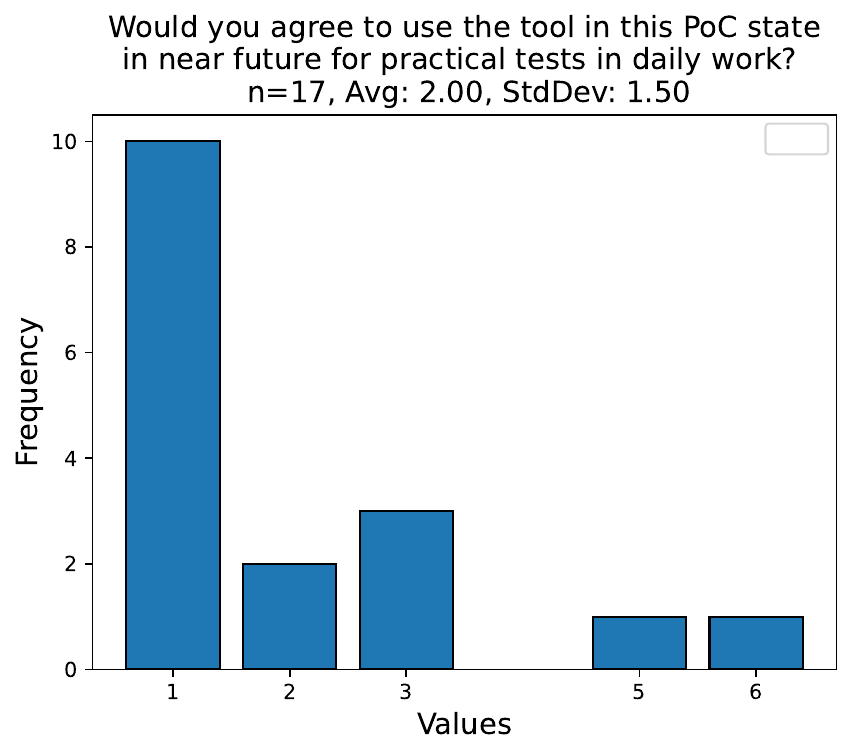}
    \end{subfigure}
  \caption{Readiness of survey participants with respect to further practical tests}\label{fig:04_automeet_readiness}
\end{figure}

The agreement to use the tooling in its \textit{proof-of-concept state in near future for practical tests in daily work} is at 2.0 on average (StdDev 1.5, n=17) on a scale from 1 (clear yes) to 6 (clear no).
33 out of 35 report that they would use a tool like \textit{AutoMeet}, considering that mentioned concerns e.g. regarding data security are addressed effectively.
The readiness of users with respect to practical tests - including the recording of meetings in general - is visualized in Figure~\ref{fig:04_automeet_readiness}.

\subsection{Requirements and constraints}
In this subsection, we list the requirements and constraints regarding a potential future \textit{AutoMeet} pipeline based on real-world user feedback.

We have collected feedback regarding four proposed privacy features. They were rated on a scale from 1 (very important/must have) to 6 (not important to me) as shown in Table~\ref{tab:04_privacy_features}.
The feature \textit{Easy starting/stopping of recording} was rated 1.76 on average (StdDev 1.48, n=38), 
\textit{Automated removal of personal data} was rated  1.87 (StdDev 1.45, n=38), \textit{Obligatory manual approval of minutes before integration to database} was rated  2.58 (StdDev 1.66, n=38), \textit{Deletion of recording after processing} was rated  2.58 (StdDev 1.73, n=38).

\begin{table}[htb]
  \centering
  %\begin{tabular}{ l c l }
  \begin{tabular}{ K{27mm} | K{55mm} X{25mm}}
    \toprule
    \textbf{Privacy feature} & 
          \textbf{Importance ranking}
          \newline 1: very important/must have \newline 6: not important to me
        & \\
    \midrule
    Option to easily start/stop recording during the meeting &
      \includegraphics[width=0.35\textwidth]{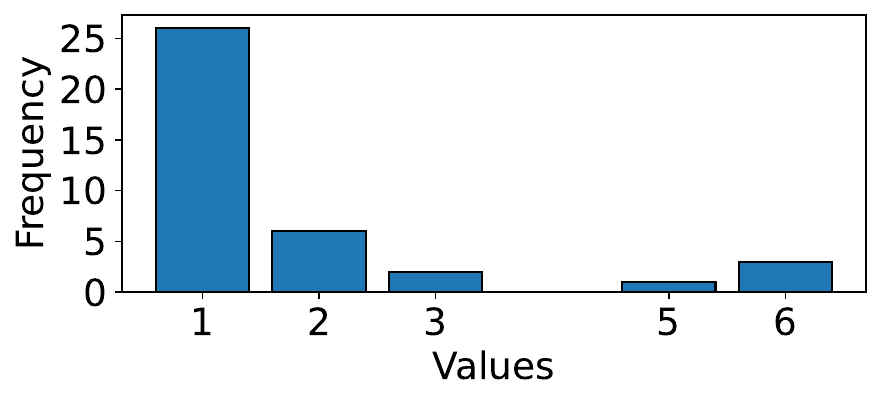} & \noindent n=38 \newline Avg: 1.76 \newline StdDev: 1.48 \\
    \midrule
    Automated removal of personal data &
      \includegraphics[width=0.35\textwidth]{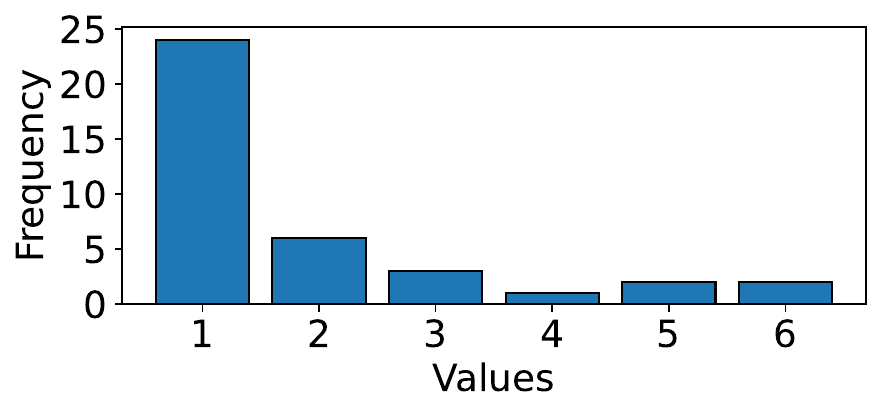} & \noindent n=38 \newline Avg: 1.87  \newline StdDev: 1.45  \\
    \midrule
    Obligatory manual approval of minutes before integration to database &
      \includegraphics[width=0.35\textwidth]{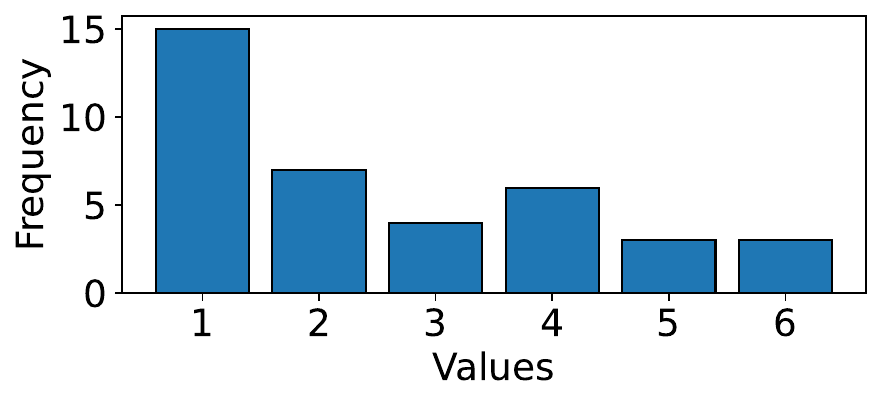} &
      \noindent n=38 \newline Avg: 2.58  \newline StdDev: 1.66  \\
    \midrule
    Deletion of recording after processing &
      \includegraphics[width=0.35\textwidth]{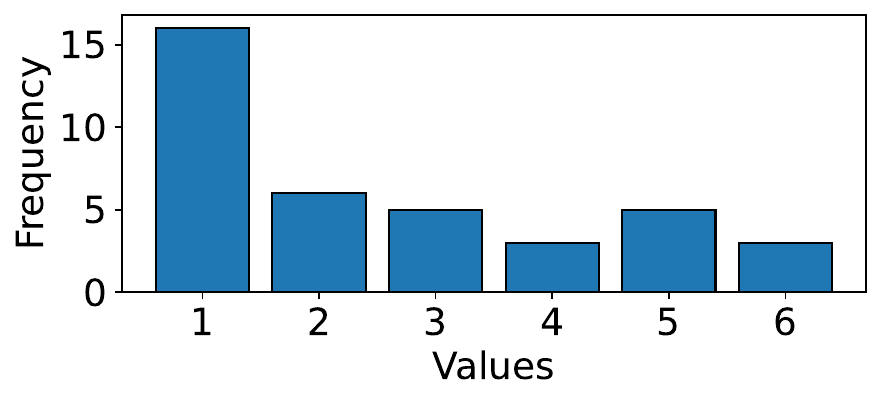} &
      \noindent n=38 \newline Avg: 2.58 \newline StdDev: 1.73  \\
    \midrule
    \bottomrule
  \end{tabular}
  \caption{Rating of four proposed privacy features by survey participants on a scale from 1 (very important) to 6 (not important)}
  \label{tab:04_privacy_features}
\end{table}

% \begin{figure}[htb]
%   \centering
%   \includegraphics[width=.9\textwidth]{04_privacy_features.png}
%   \caption{Rated importance of proposed privacy features}\label{fig:04_privacy_features}
% \end{figure}

Concerns were most frequently raised with regard to data security. Survey participants propose the following features to address these concerns: \nopagebreak[4]
\begin{itemize}
\item Secure removal of personal data and a clear concept for data protection
\item No storage of the full recordings and instead a limitation of storage solely to the agreed/edited information
\item Manual check of meeting minutes by the user before uploading to the database
\item Possibility to integrate manual meeting minutes, including minutes from offline meetings
\item Option to start/stop the recording of meetings
\item Possibility to delete meetings from the central database
\item Transparency of the underlying IT systems, especially regarding the deletion of information
\item Practical demonstrations of privacy features to build trust
\end{itemize}

Frequently proposed functional features are:
\nopagebreak[4]
\begin{itemize}
\item A search functionality to efficiently access the stored meeting minutes and preferably information from further documentation platforms
\item Access right management: the chatbot may only answer based on the information that the respective user may access.
\item Management of outdated information
\item Documentation of the hardness/quality of information along with the information itself. For example, it needs to be distinguished if a statement comes from an intern or from a senior expert or if a business statement comes from a technical person and vice versa.
\item Summarization in real-time to enable a discussion and correction of the summary during a still ongoing meeting.
\end{itemize}

%%%%%%%%%%%%%%%%%%%%%%%%%%%%%%%%%%%%%%%%%%%%%%%%%%%%%%%%%%%%%%%%%%%%%%%%%%%%%%%%%%%%%%%%%%%
%%%%%%%%%%%%%%%%%%%%%%%%%%%%%%%%%%%%%%%%%%%%%%%%%%%%%%%%%%%%%%%%%%%%%%%%%%%%%%%%%%%%%%%%%%%
\section{Discussion}
The user feedback validates our initial assumption about information being shared mainly via meetings.
Furthermore, it clearly validates the high potential of the proposed automated creation of meeting minutes along with an easy access to previous meeting minutes.
While the feedback was quite diverse and ranging from statements such as \textquotedblleft we need this as soon as possible\textquotedblright ~to \textquotedblleft I will not record my meetings\textquotedblright, users are positive about the overall approach.
Features around the protection of personal data and the option to permanently remove recorded data - among others - have been identified to be crucial for a successful implementation.
The high potential for cost savings in combination with the high sensitivity of aforementioned issues justifies corresponding investments in a high quality software system that safely prevents any misuse beyond the intended purpose.

The GPT4o model has limitations, but future LLM models can be used as drop-in replacement without any changes to the remaining pipeline.
Users that were briefed on potential errors (e.g. hallucinations) approached the automatically created summaries more positively and showed a higher readiness to correct small mistakes.
Users that were not briefed were disappointed with errors quicker.
Thus, the limitations of currently available genAI models are mainly a question of user acceptance.
This can be improved by enhancing the AI literacy of users before introducing them to the genAI tooling.
This is in line with the recommendations of \cite{HakiEtAl2025, SöllnerEtAl2025}.

Recording the meetings was identified to be difficult to implement in daily practice. The possibility to include manual meeting minutes resolves reservations of participants who are generally sceptical with regard to the end-to-end pipeline. The summarization tool could be offered on a voluntary basis to support the creation of manual meeting minutes, i.e. as a guide for the discussed topics.
A feasible option would also be to use the recording only for presentations, while relying on manual minutes for discussions and merging both the automated and manual minutes in a central database along with respective metadata.

To protect confidential data and prevent misuse, the full recordings have to be deleted permanently and securely. This would effectively address major concerns of test users. On a technical level, this could be implemented by using open source software being executed locally on user devices for the recording and summarization. Only the manually edited meetings minutes are then transferred to a central platform.

The effort to create our proof-of-concept tooling was reasonable.
The availability of open source frameworks greatly simplified the implementation.
Basically, we only needed to build a user interface around a state-of-the-art LLM model.
Such solutions are quick and easy to implement.
However our study shows, that this lacks a significant portion of the actually needed features.
For example, a database with access rights defined on the level of individual users and an effective management of the hardness/quality of information require substantial implementation efforts.
The list of features that we identify based on real-world user feedback can serve as a guideline to build respective IT systems.

Our proof-of-concept tooling is limited for the usage in this proof-of-concept setting.
We explicitly do not want to create or offer an IT system that can be deployed for long-term daily usage.
The tooling was implemented very quickly and covers the functionality in a way to assess user needs.
This is done to motivate a long-term stable solution and to identify the necessary features with a low initial implementation effort.
The approach is currently being transferred towards such a more stable deployment setup.
This will enable further insights from real-world application.

The analysis of the utilized information flow shows mainly a use of unstructured channels such as meetings and e-mails. This underlines the high potential for a centralized, well managed and easily searchable documentation platform. The reported low usage of structured information channels motivates to re-think current standards regarding documentation and information sharing beyond meetings. This could be implemented by extending the presented \textit{AutoMeet} approach with respect to the utilized database, i.e. going from a rather simple document storage towards a more sophisticated storage concept.

The feedback from our survey also underlines the high potential financial benefits of implementing an end-to-end \textit{AutoMeet} pipeline in daily practice, while highlighting the importance of considering the previously introduced privacy features.

%%%%%%%%%%%%%%%%%%%%%%%%%%%%%%%%%%%%%%%%%%%%%%%%%%%%%%%%%%%%%%%%%%%%%%%%%%%%%%%%%%%%%%%%%%%
%%%%%%%%%%%%%%%%%%%%%%%%%%%%%%%%%%%%%%%%%%%%%%%%%%%%%%%%%%%%%%%%%%%%%%%%%%%%%%%%%%%%%%%%%%%
\section{Conclusion}
We present the end-to-end approach \textit{AutoMeet} to make information exchange more efficient with the support of genAI models.
It is mainly centered around the creation of meeting minutes by a genAI model and a central documentation platform accessible via a genAI-based chatbot.
We implement a proof-of-concept tooling to assess the chances and risks in a realistic scenario.
Based on real-world feedback from potential users, we quantify the potential of such a solution and highlight features which a necessary for a successful implementation into daily work.
We see the early identification of user needs - especially regarding the requested features around data security - as an ethical and sustainable approach towards genAI-supported workflows.
We show that information is mainly shared via meetings and that \textit{AutoMeet} can make exactly this kind of information exchange more efficient.

\printbibliography

\end{document}